\documentclass[preprint,5p,times]{article}

\usepackage{xparse}%
\usepackage{morewrites}
\usepackage{wrapfig}
\usepackage{xkeyval}%

\newwrite\authorbibfile%

\AtBeginDocument{%
  \immediate\openout\authorbibfile=\jobname.aub%
}%

\AtEndDocument{%
\immediate\closeout\authorbibfile
\InputIfFileExists{\jobname.aub}{}{}
}%

\makeatletter

\define@key{authorbib}{scale}[1]{%
\def\AuthorbibKVMacroScale{#1}%
}

\define@key{authorbib}{wraplines}[10]{%
\def\AuthorbibKVMacroWraplines{#1}%
}

\define@key{authorbib}{imagewidth}[4cm]{%
\def\AuthorbibKVMacroImagewidth{#1}%
}

\define@key{authorbib}{overhang}[10pt]{%
\def\AuthorbibKVMacroOverhang{#1}%
}

\define@key{authorbib}{imagepos}[l]{%
\def\AuthorbibKVMacroImagepos{#1}%
}

\makeatother

\presetkeys{authorbib}{imagepos=l,imagewidth=4cm,wraplines=15,overhang=20pt}{}

\newlength{\AuthorbibTopSkip}
\newlength{\AuthorbibBottomSkip}
\NewDocumentCommand{\authorbibliography}{+o+m+m+m}{%
  \IfNoValueTF{#1}{%
  }{%
    \setkeys{authorbib}{#1}%
    \immediate\write\authorbibfile{%
      \string\begin{wrapfigure}[\AuthorbibKVMacroWraplines]{\AuthorbibKVMacroImagepos}[\AuthorbibKVMacroOverhang]{\AuthorbibKVMacroImagewidth}^^J
        \string\includegraphics[scale=\AuthorbibKVMacroScale]{#2}^^J
        \string\end{wrapfigure}^^J
    }%
  }%
  \IfNoValueTF{#3}{%
    \typeout{Warning: No author name}%
  }{%
    \immediate\write\authorbibfile{%
      \unexpanded{\vspace{\AuthorbibTopSkip}}^^J
      \string\noindent\relax
      \unexpanded{\textbf{#3}\par}^^J
      \string\noindent\relax
      \unexpanded{#4}^^J%
      \unexpanded{\vspace{\AuthorbibBottomSkip}}^^J
      }%
  }%
}%

\usepackage{ecrc}

\volume{00}

\firstpage{1}

\journalname{Future Generation Computer Systems}

\runauth{D. Buchaca et al.}

\jid{FGCS}

\jnltitlelogo{FGCS}

\usepackage{amssymb}

\usepackage{booktabs} 
\usepackage{subcaption}
\usepackage{algorithm}
\usepackage{algpseudocode}

\usepackage{url}

\usepackage{breakurl}
\usepackage[breaklinks,colorlinks=black,hidelinks]{hyperref}

\usepackage{amsmath}
\usepackage{bm}
\usepackage{amsfonts}  
\usepackage{color}
\usepackage{url}
\usepackage{verbatim}
\usepackage{booktabs}
\usepackage{tabularx}

\usepackage{soul}
\usepackage{ulem}

\newcommand{\old}[1]{}

\begin{document}

\begin{frontmatter}




\title{Sequence-to-sequence models for workload interference prediction on batch processing datacenters.}


\author[BSC,UPC]{David Buchaca\corref{cor1}}
\ead{david.buchaca@bsc.es}
\author[UPC]{Joan Marcual }
\ead{joan.marcual93@gmail.com}
\author[BSC,UPC]{Josep LLuis Berral}
\ead{josep.berral@bsc.es}
\author[BSC,UPC]{David Carrera}
\ead{david.carrera@bsc.es}

\address[BSC]{Barcelona Supercomputing Center (BSC),
C. Jordi Girona 1-3, 08034, Barcelona, Spain}
\address[UPC]{Universitat Polit\`{e}cnica de Catalunya (UPC) - BarcelonaTECH, Spain}

\cortext[cor1]{Corresponding author}

\begin{abstract}
Co-scheduling of jobs in data centers is a challenging scenario  where jobs can compete for resources,  leading to severe slowdowns or failed executions. Efficient job placement on environments where resources are shared requires awareness on how jobs interfere during execution, to go far beyond ineffective resource overbooking techniques.

Current techniques, most of which already involve machine learning and job modeling, are based on workload behavior summarization over time, rather than focusing on effective job requirements at each instant of the execution.
In this work, we propose a methodology for modeling co-scheduling of jobs on data centers, based on their behavior towards resources and execution time and using sequence-to-sequence models based on recurrent neural networks. The goal is to forecast co-executed jobs footprint on resources throughout their execution time, from the profile shown by the individual jobs, in order to enhance resource manager and scheduler placement decisions.


The methods presented herein are validated by using High Performance Computing benchmarks based on different frameworks (such as Hadoop and Spark) and applications (CPU bound, IO bound, machine learning, SQL queries...). Experiments show that the model can correctly identify the resource usage trends from previously seen and even unseen co-scheduled jobs.
\end{abstract}

\begin{keyword}

Resource Management \sep Sequence-to-sequence \sep Workload interference \sep Deep Learning \sep Workload placement

\end{keyword}

\end{frontmatter}



\section{Introduction}
\label{sec:introduction}

Resource under-utilization is one of the major problems in data center management, since the average utilization is estimated to be below 50\%~\cite{Reiss2012}\cite{Case2007} due to over-conservative scheduling policies, to ensure good Quality of Service (QoS) levels. However, most applications do not use resources continuously, even when there is full quota on CPU, memory and I/O to avoid degrading the QoS or violating Service level Agreements (SLAs). Flexible policies allow resource sharing among applications, while at the same time risking concurrent applications to top their requirements. Sharing resources between applications in data centers is crucial to achieve efficient utilization of those resources, reducing power consumption, allowing proper scaling out for currently running applications or accepting new applications into the system.

The principal problem when co-locating resource-sharing applications is to ensure that competition will not ruin their QoS, even when a certain tolerable degradation is expected.  Co-located applications do not need to behave as ``the sum of their behaviors'', since  interference creates a new characteristic footprint for each set of concurrent applications. Applications can be profiled in isolation in order to characterize the requirements claimed during their execution, often regarded as “black boxes” when the customer (owner of the application) does not grant access inside the application container or Virtual Machine. 
However, profiling pairs of applications in order to discover their resulting footprint becomes unfeasible, since the number of pairs grows with the number of applications in the system to be tested, which makes matters even worse as the group of co-located applications grow. Thus, there is the need to predict application resource demands and interference without the burden of running all possible combinations of applications.

The statistical analysis of monitored metrics is fundamental when automating this workload placement. Since resources are finite, smart resource sharing is encouraged by administrators in order to increase resource availability while maintaining energy efficiency\cite{Berral2010}.  Different  approaches exist for predicting resource demand and interference (slow-down produced by resource competition) on co-located applications. 
Nevertheless, most methods are based on predicting overall interference. When applications show differentiated phases of behavior over time, aggregation cannot capture the true nature of resource demand, thereby giving rise to suboptimal scheduling decisions. Most classic machine learning approaches for this problem, such as ~\cite{Delimitrou2013a} or \cite{Mishra2017},   do  not consider the temporal dimension of executions, and therefore do not predict resource usage over time. This means that schedulers are obliged to optimize fixed blocks of expected resource usage. In this scenario, schedulers can take suboptimal decisions to block possible colocations of applications that only complete for a fraction of their execution. 


In this work we present ResourceNet, a workload-to-workload forecasting methodology to predict the effects of application co-location interference, using sequence-to-sequence models based on Recurrent Neural Networks (RNNs). RNNs are powerful models with the capability of dealing with time series. Of equal importance, RNNs are able to deal with inputs and outputs of arbitrary lengths. The method presented herein predicts the footprint for resource demands of co-located applications, given that the traces of these applications run in isolation. Furthermore, we show how this model can be used to predict accurately the run-time of applications sharing resources.

Our model employs two Gated Recurrent Units (\textit{GRU}) as building blocks; one \textit{GRU} processes the trace signal of the incoming applications and passes the processed information to the other \textit{GRU}, which outputs the expected resources of the co-located applications over time. The model predicts the whole resource demand trace throughout execution, thereby providing schedulers with a sufficiently accurate estimation for placing applications together and thus minimizing interference. Training the model using a diverse set of benchmarking applications enables it to attempt predictions to unseen co-located applications. The recurrent nature of the model allows it to process input sequences or different (and arbitrary) lengths. Moreover, it is able to generate output sequences of arbitrary length, thereby making our solution adaptable enough to address the problem in question.


We validate the proposed methodology by computing the error of the predicted resources (of co-joined application traces) with respect to the real resources. We use benchmarks from Big Data applications (from both Apache Hadoop and Apache Spark) as workload data. We have selected Hadoop and Spark benchmarking workloads because of their popularity in High Performance Computing (HPC) applications, where scheduling and environment configuration have a high impact on the application performance. Experiments measure the prediction of the low-level resource usage (i.e. CPU, memory and I/O) of pairs of co-located workloads over time. The different benchmarks we use belong to the Intel HiBench~\cite{HiBench}, the IBM SparkBench~\cite{Li:2015:SCB:2742854.2747283} and the Databricks Spark-Perf benchmarks~\cite{sparkperf}. The method is trained and evaluated on low-level monitoring metrics which are reasonably available on any HPC setting. Our method is compared against different simpler machine learning alternatives in order to assess the possible drawbacks of using standard models. 

For our experiments, we created a dataset with 
 execution traces from the previously mentioned benchmark suites. The dataset consists of triplets $(\bm{a}, \bm{b}, \bm{a} \wedge \bm{b})$ where $\bm{a}$ and $\bm{b}$ contain the traces of the isolated executions from two applications,   and $\bm{a} \wedge \bm{b}$ represents an execution of the co-located pair. 
 To build a reasonable dataset, we generated scenarios in which both appli- cations did not start at the same time. In particular, we created co-scheduling situations in which the start time for one of the applications was shifted by a factor of 25\%, 50\% and 75\% of the length of the longest application being co-scheduled.

\section*{Contributions }

The main contributions of this paper can be summarized as follows:
\begin{itemize}
\item A novel application of Recurrent Neural Networks that translates the characterization of two applications  $\bm{a}$ and  $\bm{b}$ in isolation, into the predicted trace of the co-execution of such applications $\bm{a} \wedge \bm{b}$. 

\item A novel feature, \textit{percentage completion time},  for estimating the completion time of co-scheduled applications. This feature improves predictions  made by using the standard stopping criteria based on the \textit{end of sequence} feature. 


\item A comprehensive evaluation of the method against other relevant machine learning approaches. We show the ad- vantages of our method, which  are  especially  noticeable in two cases: when co-located executions have different lengths, and when co-scheduling heavily impacts the exe- cution time of the applications due to high interference.
\end{itemize}

The proposed method can be used to characterize applications according to their compatibility with other workloads. Furthermore, it enables resource managers to plan resource allocation and load balancing better in advance.

The rest of the paper is structured as follows:  Section~\ref{sec:background}  provides background on the commonly used recurrent neural network methods. Section~\ref{sec:methodology} presents the methodology, scenario and used techniques. Section~\ref{sec:dataset} provides a description of the data used  for experimentation. Section~\ref{sec:experiments}  describes the experiments conducted to validate this work. Section~\ref{sec:relatedwork} describes  Related Work, and finally Section~\ref{sec:conclusions} discusses the conclusions and future work.

\section{Background on RNNs}
\label{sec:background}

In this section we present an overview of the fundamentals on Recurrent Neural Networks (RNNs) and some important enhancements on the standard models. We also provide the building blocks of our model. 
For the rest of this work, we denote $\bm{x}_t$ a vector of $d$ components (containing monitored metrics) at time step $t$. A sequence of $n$ vectors  are denoted as $\bm{x}_{1:n}$. The $k$'th coordinate from a vector $\bm{x}_t$  is denoted as $\bm{x}_t[k]$. 


\subsection{Recurrent Neural Networks}

A Recurrent Neural Network  is a function that maps a sequence of input vectors into an output vector. Specifically, the RNN takes a sequence $\bm{x}_{1:t}$, where each $\bm{x}_j \in \mathbb{R}^{d_{in}}$, and returns a single output vector $\bm{y}_t \in \mathbb{R}^{d_{out}}$, depending on a set of parameters $\bm{\theta}$ that need to be learned. Equation~(\ref{eq:RNN_forward_a}) denotes the output of the RNN at time $t$, while equation~(\ref{eq:RNN_forward_b}) denotes the whole sequence of outputs $\bm{y}_{1:n}$ produced by applying (\ref{eq:RNN_forward_a}) at every time step and concatenating the output results.

\begin{subequations}
\label{eq:RNN_forward}
\begin{align}
\label{eq:RNN_forward_a}
	\bm{y}_t = \text{RNN}(\bm{x}_{1:t}; \bm{\theta}) \\
\label{eq:RNN_forward_b}
	\bm{y}_{1:n} = \text{RNN}^*(\bm{x}_{1:n}; \bm{\theta}) 	
\end{align}
\end{subequations}

Output $\bm{y}_t$ is used to predict relevant information about the problem at hand at time $t$. Since we use the RNN to predict the expected resource demand of two concurrent  applications at time $t$, $\bm{y}_t$ will be a vector containing the resources used by the two co-located applications.

\begin{figure}[!h]
\centering
  \includegraphics[width=0.5\linewidth]{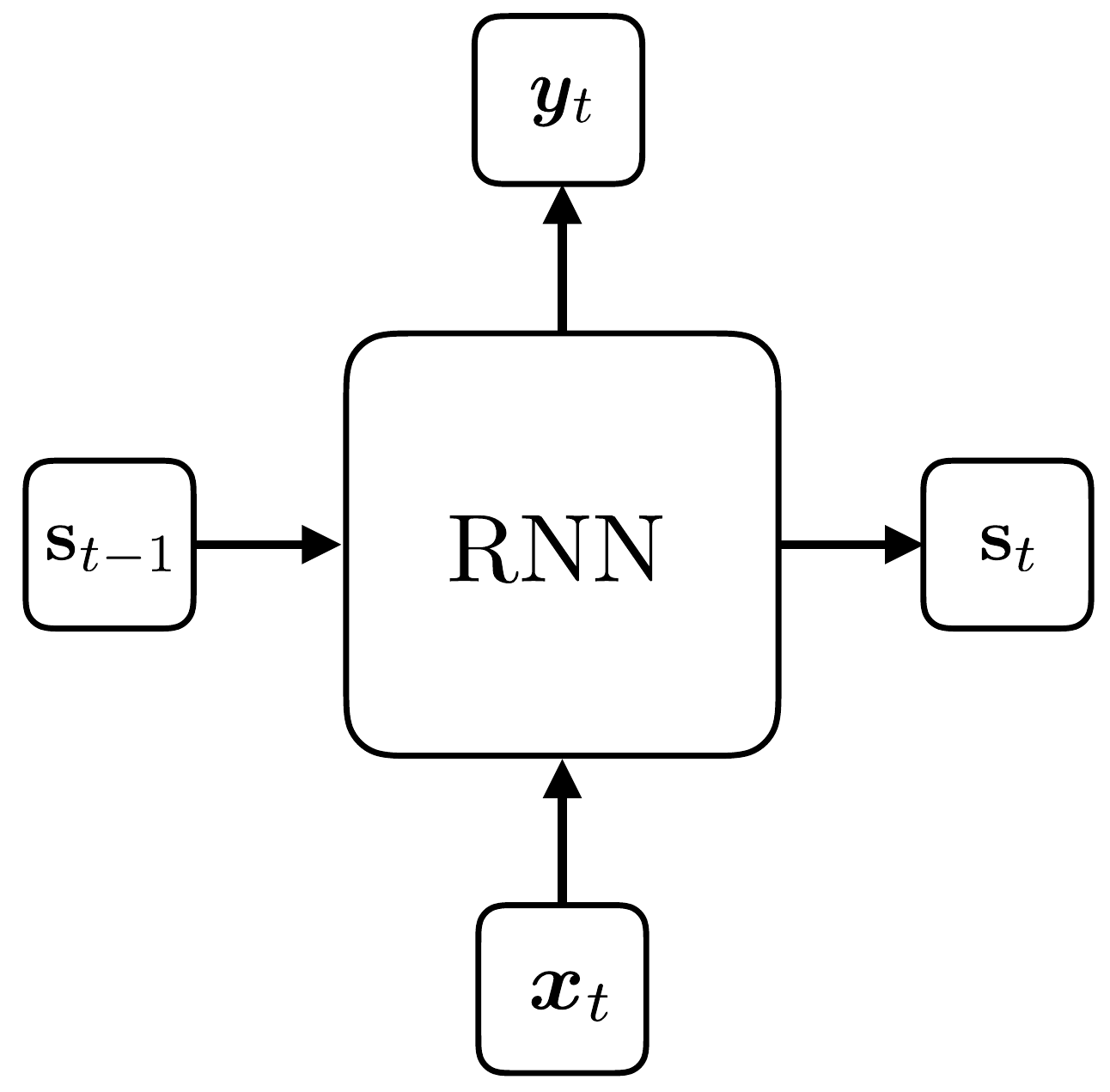}
 \caption{The computation of $\bm{y}_t$ and $\bm{s}_t$ depends on  $\bm{s}_{t-1}$ and $\bm{x}_t$.  }
  \label{fig:RNN_cell}
\end{figure}

In order to keep information from previously seen elements in sequences, RNNs have a hidden state $\bm{s}_t$ that is updated at every time step. The hidden state influences the predictions of the model over time. 
Although the hidden state is implicit in the equations~(\ref{eq:RNN_forward_a}) and~(\ref{eq:RNN_forward_b}), we can make the state explicit by using the notation $(\bm{y}_t, \bm{s}_t) = \text{RNN}(\bm{x}_t, \bm{s}_{t-1}; \bm{\theta})$. This indicates that the output $\bm{y}_t$ has been computed through input $\bm{x}_t$ and previous state $\bm{s}_{t-1}$. Figure~\ref{fig:RNN_cell} shows a diagram of an RNN.

Notice that while $\bm{y}_t$  (the output at time $t$) is not affecting $\bm{y}_{t+1}$, the updated state $\bm{s}_t$ will affect next state $\bm{s}_{t+1}$. The initial state $\bm{s}_0$ is usually defined as a vector of zeros, unless some other known contextual information can be provided for the model.  For example, in \textit{sequence-to-sequence} models (described in Section~\ref{seq2seq}) the initial hidden state of the decoder plays the crucial role of  providing a summary of the input sequence to the decoder.
The way  in which $\bm{s}_t$ is computed is what mostly differentiates RNN models, such as  Gated Recurrent Units (GRU) and Long Short-Term Memory (LSTM). Independently of how the hidden state is computed, RNNs are usually trained  by using optimization algorithms based on  Stochastic Gradient Descent (SGD) in order  to minimize the difference between the predicted outputs and the expected outputs. 


\subsection{Gated Recurrent Unit}
 \label{subsection:GRU}

Gated Recurrent Units (GRU) are a type of recurrent neural network designed to avoid gradient vanishing problems~\cite{Cho2014}. The GRU is one of the simplest RNNs employing a gating mechanism that enables the network to learn which information needs to be kept over time and which can be forgotten. Equations~(\ref{z_GRU}) to~(\ref{s_GRU}) define, at time $t$, the output state $\bm{s}_t$, given the input data $\bm{x}_t$ and the previous state $\bm{s}_{t-1}$. In the presented equations, $\sigma$ is the sigmoid function.
\begin{subequations}
\label{GRU_eq}
\begin{align}
\label{z_GRU}
\bm{z}_t			&= \sigma \left( \bm{W}^{xz} \bm{x}_t + \bm{W}^{sz} \bm{s}_{t-1} + \bm{b}^z \right) \\ 
\label{r_GRU}
\bm{r}_t			&= \sigma \left( \bm{W}^{xr} \bm{x}_t + \bm{W}^{sr} \bm{s}_{t-1} + \bm{b}^r \right) \\
\label{s_tilde_GRU}
\tilde{\bm{s}}_t	&= \tanh  \left( \bm{W}^{xs} \bm{x}_t + \bm{W}^{ss} (\bm{r}_t \odot \bm{s}_{t-1} ) + \bm{b}^s \right) \\
\label{s_GRU}
\bm{s}_t			&= \tilde{\bm{s}}_t \odot \bm{z}_t    + (1 - \bm{z}_t) \odot \bm{s}_{t-1} 
\end{align}
\end{subequations}


The output state $\bm{s}_t$ is controlled by two vectors: the update gate $\bm{z}_t$ and the reset gate $\bm{r}_t$  defined in equations~(\ref{z_GRU}) and~(\ref{r_GRU}) respectively. Since both vectors come from applying a sigmoid function, most of their components will be close to 0 or 1, controlling the information passed by any element-wise multiplied vector. In extreme cases, when the gate components are 0 or 1, the flow of information can be completely blocked or remain intact. For example, if  the reset gate $\bm{r}_t$ is zero and the update $\bm{z}_t$ is one, the previous hidden state $\bm{s}_{t-1}$ does not affect the next state $\bm{s}_{t}$. 

Once the reset and update gates are computed, a proposed hidden state  $\tilde{\bm{s}}_t $  is created. Equation~(\ref{s_tilde_GRU}) defines the proposed hidden state $\tilde{\bm{s}}_t $ at time $t$. Notice that $\tilde{\bm{s}}_t $ depends on the input vector and the information from the past hidden state $\bm{s}_{t-1}$ that is allowed to pass by the reset gate $\bm{r}_{t}$. Afterwards the true state $\bm{s}_t$ is computed as a combination of components from the proposed state $\tilde{\bm{s}}_t$ and the previous state $\bm{s}_{t-1}$. Notice that in equation~(\ref{s_GRU})  the element-wise multiplications $ \tilde{\bm{s}}_t \odot \bm{z}_t $  and $ (1 - \bm{z}_t) \odot \bm{s}_{t-1} $  can control which components of the proposed state and the previous state are kept. The matrices $\bm{W}^{\ast \ast}$ and vectors $\bm{b}^\ast$ are the weights adjusted during learning. 


\subsection{Sequence-to-sequence models}
\label{seq2seq}

Most neural network architectures prior to~\cite{Sutskever2014} were designed for problems where the goal is to map input vectors $\bm{x}$ to fixed size output vectors $\bm{y}$. Sequence-to-sequence models proposed an end-to-end trainable neural network architecture designed to tackle sequential data. The architecture is based on two Recurrent Neural Networks named \textit{encoder} and \textit{decoder}. The encoder is designed to read the input sequence and produce a vector that is passed to the decoder. Since the hidden state of the encoder is updated by reading all the information from the input, it can be interpreted as a summary of the input sequence. The decoder reads the encoded vector and produces a new sequence, but instead of initializing the hidden state of the decoder with zeros, it is initialized to the last hidden state of the encoder RNN. These types of models have been successfully applied to different tasks ranging from machine translation, e.g~\cite{Sutskever2014}\cite{Bahdanau2014}, to image captioning~\cite{Vinyals2015}.

\subsection{Attention mechanism}

The main drawback of  the \textit{sequence-to-sequence} approach explained in Section~\ref{seq2seq} is that the whole input sequence is compressed into a single vector. 
Since the amount of parameters and the computational cost of running RNNs grows with the number of neurons in the hidden state, it is not feasible to use hidden states sizes of the order of $n \cdot d$, where $n$ is the number of elements in the sequence and $d$ the number of features. Storing all the information of a sequence into a single vector of fixed size is therefore a difficult task.

The work done by Bahdanau et al.~\cite{Bahdanau2014} improves the architecture of the previously mentioned sequence-to-sequence model by furnishing the decoder with an \textit{attention mechanism}. At every step of the decoding process, this mechanism provides a vector containing information about what \textit{might} be relevant from the input sequence. Therefore, the decoding process is not performed from a single vector as in~\cite{Sutskever2014}, but from a vector generated  at every time step. This vector generated by the \textit{attention mechanism} is called the context vector. In ~\cite{Bahdanau2014} authors showed that the performance of  a sequentence-to-sequence model with attention did not degrade for sequences of various lengths, whereas the same model without attention gave far worse results for long sequences.

The decoding process for an RNN with attention works as follows: First, the encoder reads the input sequence and generates the matrix $\bm{E}=~\text{RNN}_\text{enc}^*(\bm{x}_{1:n}; \bm{\theta})$. Then, at time $t$, the decoder $\text{RNN}_\text{dec}$ receives as input (in addition to any recurrent inputs)  the concatenation of two vectors: $[\bm{\tilde{y}}_{t-1} ; \bm{c}_t ]$.  This concatenation provides extra information at the decoder at every time step. 
The \textit{context} vector  presented in~\cite{Bahdanau2014} is defined by equations~(\ref{mu_eq}) to~(\ref{context_eq}). 

\begin{subequations}
\label{attention_eqs}
\begin{align}
\label{mu_eq}
\bm{\mu}_t [k]		=& \bm{v}^T \tanh(\bm{W}^{\bm{\mu}}\bm{E}[\bm{\cdot}, k] + \bm{V}^{\bm{\mu}} \bm{s}_{t-1} + \bm{b}^{\bm{\mu}}) \\
\label{alpha_eq}
\bm{\alpha}_t	       =& \text{softmax} ( \bm{\mu}_t ) \\
\label{context_eq}
\bm{c}_t		            =& \bm{E} \bm{\alpha}_t
\end{align}
\end{subequations}

  In  (\ref{context_eq})  the  \textit{context} vector is computed as a matrix vector product, which can be interpreted as a  weighted sum of the columns of  $\bm{E}$. The weights given to the columns of $\bm{E}$ are the  values of the \textit{attention} vector $\bm{\alpha}_t$. 
        The vector $\bm{\alpha}_t$ from  (\ref{alpha_eq}) is called the \textit{attention} vector because it contains the weights used to ``focus'' (or attend) to different parts of $\bm{E}$. This vector is computed by normalizing (with a softmax function)  the \textit{attention} energy $\bm{\mu}_t$.  The  \textit{attention} energy is computed in~(\ref{mu_eq}) by using a single  Multi Layer Perceptron (MLP), with a single  tanh layer. The MLP predicts a single scalar which defines the  $k$'th coordinate from $\bm{\mu}_t$.  The MLP is applied at every position $k$  from 1 to $n$, where $n$ is the number of columns of $\bm{E}$, making  $\bm{\mu}_t , \bm{\alpha}_t \in \mathbb{R}^{n \times 1}$.  
  Notice that the product of $\bm{W}^{\bm{\mu}} \bm{E}[\bm{\cdot}, k]$ in~(\ref{mu_eq}) does not depend on the time step $t$  and therefore it can be precomputed in advance.  

One of the key properties of the presented  \textit{attention}  mechanism is the ability to construct a fixed-size  vector $\bm{c}_t$ of dimension $d$ which does not depend on the length of the input sequence. The vector $\bm{\mu}_t$  has variable size because, for a given sequence of arbitrary length $n$, vector $\bm{\mu}_t$ is computed by applying  (\ref{mu_eq}) for $k \in \{1,\dots,n\}$. Nevertheless, the  product  $\bm{E}\bm{\alpha}_t$  is  a vector of size $d$, where $d$ is the number of rows in $\bm{E}$ because  $\bm{E} \in \mathbb{R}^{d \times n}$  and therefore $\bm{c}_t = \bm{E} \bm{\alpha}_t \in \mathbb{R}^{d \times 1}$.




In many applications, $\bm{E}$ is generated using a bidirectional RNN which simply concatenates the hidden states of two RNNs. One $\text{RNN}_\text{f}$  receives the input sequence forwards (from $x_1$ to $x_n$) while the other  $\text{RNN}_\text{b}$  receives the sequence backwards (from $x_n$ to $x_1$). In~\cite{Bahdanau2014}  
  $\bm{E}$ is created by stacking $\text{RNN}_\text{f}^*(x_{1:n})$ on top of $\text{RNN}_\text{b}^*(x_{n:1})$. 

\section{ResourceNet}

\label{sec:methodology}

ResourceNet is a fully learnable system based on a sequence-to-sequence architecture with an attention mechanism. 
The model  takes as input pairs of sequences containing measurements of applications run on isolation. The input sequences are aggregated and fed to a decoder which  produces the expected measurements  of running both co-located applications  sharing resources.  
Since the model is based on a sequence-to-sequence architecture it can naturally work with sequences of different lengths, which makes it suitable for the scenario presented here, in which measurements belong to the workload traces.



\begin{figure}[!h]
\centering
  \includegraphics[width=1.04\linewidth]{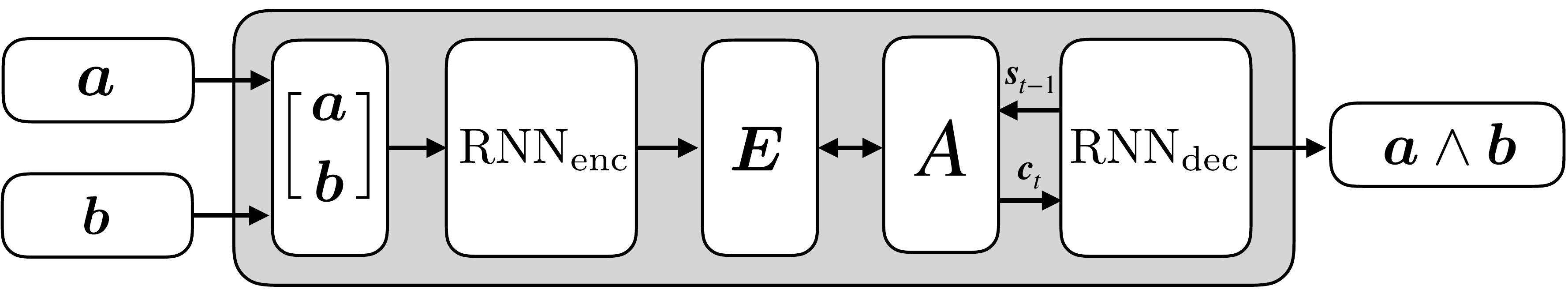}
  \caption{Input-output diagram of the proposed model}
  \label{fig:ResourceNetDiagram}
\end{figure}

Figure~\ref{fig:ResourceNetDiagram} shows the diagram for the model, indicating the inputs and outputs, and the four building blogs composing the data pipeline. The entire shaded box represents the whole model at a high level. The first component, $\left[\bm{a}\atop \bm{b}\right]$, performs the joining operation where input sequences $\bm{a}$ and $\bm{b}$ are stacked to form a single matrix of measurements; this process is explained in detail in Subsection~\ref{subsection:stacking}. The second component, $\text{RNN}_\text{enc}$, encodes the input sequences with a GRU and  produces as output a matrix $\bm{E}$. The third component, $\bm{E}$, represents the  stored  encoded input sequences. The fourth component, $\bm{A}$ is the attention mechanism, which receives the hidden state from the decoder and generates a context vector that is fed to $\text{RNN}_\text{dec}$. The last component of the diagram is the decoder. The decoder is  a GRU that generates as output the expected resource demands of the co-scheduled applications. We  denote those resource predictions by $\bm{a} \wedge \bm{b}$.  

The learning process is performed by minimizing the Mean Squared Error loss function, which computes the error between the predicted outputs and the true trace values at every time step. We change the parameters of the model using a  gradient descent procedure.

\subsection{Stacking measurements of the input sequences}
\label{subsection:stacking}

The input sequences of the model are stacked in order to generate a single matrix of measurements. Given two vectors $\bm{v}=(v_1,\dots, v_p)^\top$ and $\bm{w}= (w_1,\dots, w_q)^\top$ we will denote by $ \left[ \bm{v} \atop \bm{w} \right]$ or $ [\bm{v}; \bm{w}]$ the vector $(v_1,\dots, v_p, w_1,\dots, w_q)^\top$.

Given two arbitrary sequences $\bm{a}$ and $\bm{b}$ of length $\text{l}(\bm{a})$ and $\text{l}(\bm{b})$ respectively we denote by $\left[ \bm{a} \atop \bm{b} \right]$ or $[\bm{a}; \bm{b}]$ a sequence of length $N= \max( \text{l}(\bm{a}), \text{l}(\bm{b}) )$ containing the measurements of both sequences at every time step. By convention we write the longest sequence first. 
Since sequences may differ in length we append the shortest sequence with zeros to match the length of the longest sequence. If we denote by $\bm{0}_d$ a vector containing $d$ zeros where d is the number of elements in $\bm{b}_k$ we have:
\begin{equation*}
\left[ \bm{a} \atop \bm{b} \right] := \left(  \left[ \bm{a}_1 \atop \bm{b}_1 \right]  , \dots, \left[ \bm{a}_{\text{l}(\bm{b}) } \atop \bm{b}_{\text{l}(\bm{b})} \right], \left[ \bm{a}_{\text{l}(\bm{b}) +1 }  \atop \bm{0}_d \right], \dots,  \left[ \bm{a}_{\text{l}(\bm{a}) }  \atop \bm{0}_d \right] \right)
\end{equation*}

$\newline$
{\bf Example}: Let us consider $\bm{a}= (\bm{a}_1, \bm{a}_2, \bm{a}_3, \bm{a}_4)$ and $\bm{b} = (\bm{b}_1, \bm{b}_2)$ containing measurements on $\mathbb{R}^2$. That is $\bm{a}_i, \bm{b}_j \in \mathbb{R}^2$ for any valid $i,j$. Then the generated matrix for two applications that start at the same time is  $ \left[ \bm{a} \atop \bm{b} \right] := \left(  \left[ \bm{a}_1 \atop \bm{b}_1 \right]  , \left[ \bm{a}_2 \atop \bm{b}_2 \right], \left[ \bm{a}_2 \atop \bm{0}_2 \right], \left[ \bm{a}_4 \atop \bm{0}_2  \right] \right) $.

 
$\newline$
The notation $\left[\bm{a}, \bm{b}\right]$ is reasonable for describing the situation in which the model is fed with traces of programs that start execution at the same time. Nevertheless, we may want to predict scenarios where applications  do not start at the same time. In order to provide the network with such information, we pad a sequence with zeros before it starts.
If we wish to tell ResourceNet that application $\bm{b}$ starts $k$ time-steps after $\bm{a}$,  we  add $k$ vectors containing zeros at the beginning of $\bm{b}$.  
In order to make ResourceNet capable of making  predictions in that environment, where applications are not required to start at the same time, our training data will contain co-scheduled applications starting with different time delays.  

\subsection{Encoding input traces, Decoding co-scheduled trace}
\label{subsection:encoder}

The encoding process reads the traces from the isolated runs and generates the matrix $\bm{E}=~\text{RNN}_\text{enc}^*(\bm{x}_{1:n}; \bm{\theta})$, using the equations of the GRU from Subsection~\ref{subsection:GRU}. To this end, we use a GRU as our encoder. 

The decoding process takes the produced $\bm{E}$ as input and generates the output sequence one vector at a time. 
This output vector has as many components as it does features we wish to predict (in our case, 9), plus some features  used to decide when applications finish.  
A detailed description of the features can be found in Section \ref{sec:dataset}.
The decoder $\text{RNN}_\text{dec}$ receives as input at time $t$ the vector $\bm{x}_t:=~\left[ \tilde{\bm{y}}_{t-1} ; \bm{c}_t \right]$ which is the concatenation of $\tilde{\bm{y}}_{t-1}$ and $\bm{c}_t$ (in addition to any recurrent inputs).
Vector $\tilde{\bm{y}}_{t-1}$ is the  output predicted by the decoder at the previous time step. Vector  $\bm{c}_t$  is the ``context vector'' generated by the attention mechanism of the decoder. Given a hidden state  $\bm{s}_t$ and a input $\bm{x}_t$  the predicted resources of the colocated applications at time $t$ is $\tilde{\bm{y}}_{t}  =  \text{RNN}_{\text{dec}}(\bm{x}_t, \bm{s}_t; \bm{\theta})$. 




\subsection{Predicting completion time}
\label{subsection:completion_time}
We have experimented with two mechanisms for predicting the completion time of the co-scheduled jobs.  The standard End of Sequence  (EOS)  feature approach and our own Percentage Completion (PC) feature approach.

\subsection*{End of Sequence feature}

The  first strategy, based on an End of Sequence feature, is the standard approach used to decide when an RNN should stop producing more vectors. This feature vector is takes value 0 at every time step except the last one, where it takes value 1. This vector simply  tells  the decoder when both applications finish.  If our decoder can predict  $\text{EOS}$  with reasonable precision then we can build stopping criteria based on those values to predict the completion time of the co-scheduled applications.

 The presented End of  Sequence  approach for our problem is not equivalent to the analogous problem in other domains such as Neural Machine Translation (NMT). In NMT, the decoder emits conditional probability over the vocabulary at every time step which can be used to select the word with the highest probability. In that scenario, $\text{EOS}$ is just a special word that stops the decoding process, which means that if the decoder emits the ``EOS'' symbol the decoding process is stopped. In our setting the EOS is a new feature that takes a real value at every time step. 


\subsection*{Percentage Completion Features}

The second strategy consists of  a novel approach 
based on  two additional features (one per job) which we call Percentage Completion (PC) features. PC features keep track of the percentage completion  of the workloads, providing ResourceNet with extra information relevant to  the job estimation runtime. 


 We  denote by  $\text{PCF}_{\bm{a}}$ and $\text{PCF}_{\bm{b}}$  the percentage completion features for input sequences $\bm{a}$ and $\bm{b}$, respectively.  Both features contain at time $t$ how much of the workload has been completed until $t$, expressed as a percentage.  For a given sequence $\bm{s}$ of length $N$, we define  $PCF_{\bm{s}} = \left( \frac{1}{N} \cdot 100 , \frac{2}{N} \cdot 100 , \dots, \frac{N}{N} \cdot 100  \right)$. Notice that, by construction, PCF features contain monotonically increasing values that  must finish with value 100.  Nevertheless the rate of increment at every time step will depend on the overall number of time steps of the sequence.

$\newline$
{\bf Example}: Let us consider $\bm{a}= (\bm{a}_1, \bm{a}_2,  \dots , \bm{a}_{100})$ and $\bm{b}=(\bm{b}_1, \bm{b}_2, \dots, \bm{b}_{300})$. Then  $\text{PCF}_{\bm{a}}=(1, 2, \dots, 100)$ and $\text{PCF}_{\bm{b}}=(0.333, 0.666, 1.0, \dots, 100)$ are two vectors of length 100 and 300 respectively.



\section{Workload data}
\label{sec:dataset}

\subsection{Monitoring running applications}

In order to capture the trace of each execution we have profiled them by using the basic Linux performance analysis tools: \textit{vmstat}, \textit{iostat}, \textit{ifstat} and \textit{perf}. These tools gather system performance metrics of running processes in a non-invasive way. Additionally, these tools have a lower performance impact in the system, causing negligible overhead to the executions. From these tools we have gathered a total of 141 features with time granularity of one second. 

For our study we have selected 9 key features, shown in Table~\ref{FeatureTable},  that are especially relevant for interference prediction and resource estimation~\cite{phases}\cite{Delimitrou2013a}\cite{Marco2017}.

\begin{table}[h]
\centering
\begin{tabularx}{\linewidth}{|l|X|}
\hline
\textbf{Feature}     & \textbf{Description}                           \\ \hline
CPU                  & Percentage of total CPU used                   \\ \hline
RAM                  & Amount of Bytes of main memory used            \\ \hline
IOR                  & Blocks received from a block device (blocks/s) \\ \hline
IOW                  & Blocks send to a block device (blocks/s)       \\ \hline
CPI                  & Cycles per instruction                         \\ \hline
LLCM                 & Last level cache misses                        \\ \hline
FLCM                 & First level cache misses                       \\ \hline
PF                   & Page faults                                    \\ \hline
TLBM                 & Translation lookaside buffer misses            \\ \hline
\end{tabularx}
\caption{Workload metrics recorded at each time step.}
\label{FeatureTable}
\end{table}
The dataset used in the experiments contains traces generated by a variety of micro-benchmarks (workloads). The workloads used have been extracted from  different suites: HiBench~\cite{HiBench}, Spark-perf~\cite{sparkperf} and SparkBench~\cite{Li:2015:SCB:2742854.2747283}. These suites contain different  Big Data workloads, and have also been used in similar studies~\cite{spark-perf-marenostrum}\cite{hibench-used}\cite{spark-perf-MLlib}\cite{spark-perf-workload-characterization}. The benchmarks include machine learning, data mining and big data benchmarks. Some examples of workloads are executions of \textit{PageRank}, \textit{K-means}, \textit{Naive Bayes},  \textit{Logistic Regression}, etc. The traces are executed using a server with two Intel Xeon E5-2630 processors and 128~GB of RAM,  up to 400  isolated and co-located executions.

\subsection{Dataset description}
\label{sec:datasets}

The dataset is composed of workloads \textit{triplets}. Each triplet contains a combination of three execution traces. 
The first two traces correspond to isolated executions of the two applications. The third trace contains the  execution of the  co-located application from the first two traces.  Therefore, the data has the form  $D~=~\left\{(\bm{w}^{[i]}, \bm{w}^{[j]}, \bm{w}^{[i]}\wedge \bm{w}^{ [j]}) \mid i,j \in I \right\}$ where $I$ is a set of indices of the workloads.

In real world scenarios, co-located applications do not necesarilly start at the same time. In order to increase the co-location cases in our collected dataset, we prepared different scenarios where co-located applications $\bm{a}$ and $\bm{b}$ start with different delays. For the benchmarking executions, one of the concurrent applications is delayed to start after its co-located peer application. The phase differences used in the dataset generation are 0 (synchronized), 0.25, 0.5 and 0.75. A phase difference of 1.0 is not taken into consideration since it would mean applications running completely in isolation, one after the other.
Notice that bigger differences usually give rise to less interference, since applications have fewer runtime sharing resources.

\section{Experiments}
\label{sec:experiments}

ResourceNet is validated through a test set of executions which are used to evaluate the prediction capabilities in different job co-location scenarios. 
We use the Mean Absolute Percentage Error (MAPE) to assess the quality of the predictions at every time step of the execution trace.  
We have designed two experiments: 


\begin{itemize}
\item 
In Experiment 1 we evaluate the quality of the predictions when both applications are running with the different models presented. We present a table with the error metrics computed in a test set.

\item 
In Experiment 2 we evaluate the accuracy of the model presented when predicting the runtime of co-scheduled applications. We present experiments with different methods to assess the job runtime of the co-scheduled applications.
\end{itemize}

\subsection*{Evaluation methodology}


To evaluate the advantages of the proposed method we compare it with other sensible alternatives. Firstly, we use a na\"ive baseline model to estimate resources. The baseline model predicts the resource usage at time $t$ as the sum of the resources of the isolated applications at time $t$.  This means that the output at time $t$ for a given input  $\left[ \bm{a} ; \bm{b} \right]$ 
  is $\tilde{\bm{y}}_t = \bm{a}_t + \bm{b}_t$. Notice that $\left[ \bm{a};  \bm{b} \right]$ is a matrix of features that already contains padded zeros to encode any temporal phase difference of sequences (should they exist). Therefore, if $\bm{b}$ starts $k$ time steps after $\bm{a}$ then this baseline should predict correctly the features values of the co-located applications for the first $k$ time-steps (since there is a single application running and there is no competition for resources).

We also compare our approach with a linear regression (LR) and a Multi-Layer Perceptron (MLP).  
 The two regression models present a natural improvement over the baseline approach. Both methods predict the resources usage at time $t$  as a function of the input features $\bm{a}_t$ and $\bm{b}_t$.  The baseline, the LR and  the MLP  take an input  vector containing measurements  from the isolated sequences at time $t$  and  estimate the features of the the co-located sequence at time $t$. This approach involves two critical problems:
\begin{itemize}
\item The temporal nature of the data suggests that values at time $t$ might be correlated with nearby values . This is not taken into consideration.
\item If the input has length $n$ and the output has length $n+k$ there will be $k$ time-steps where the models cannot make predictions, 
because the model has no input data from which to make predictions.
\end{itemize}
The first problem can be somewhat mitigated  by using a sliding window mechanism over the input data. Nevertheless, 
including a sliding window mechanism increases the complexity of the solution and adds a new hyperparameter  to be tuned (the length of the sliding window).

 The second problem is even harder to assess. A sensible approach  could be to pad  the execution with zeros in the $k$ time steps where there is no data. However,  this solution invents timesteps with no resource usage and does  not address the issue that, in reality,  co-scheduled executions resource consumption ``stretches in time'' when co-scheduled jobs share resources. 
 Models therefore require some sort of  memory from past values in order to understand the effect left produced by the co-location over time. 
 Sequence-to-sequence models with an attention mechanism naturally deal with these two issues.


\subsection{Experiment 1:  Resource usage predictions}

In this experiment we evaluate  the accuracy of the predictions made by the different models on co-scheduled jobs.
Table~\ref{tab:ResultsTable} contains the MAPE errors of the predictions on the test set.
The best results are obtained by ResourceNet, followed by the Multilayer perceptron. Notice that for some metrics, such as CPU and IOR, the error  produced by the proposed model is reduced more than half when compared to the other models. Moreover, the model presents a lower  standard deviation in most cases. 

\begin{table*}[h!tb]
\normalsize
\centering
\begin{footnotesize}
\begin{tabularx}{\textwidth}{X rr rr rr rr}
\hline
  & \multicolumn{2}{c}{baseline} & \multicolumn{2}{c}{linear regression} & \multicolumn{2}{c}{multilayer perceptron} & \multicolumn{2}{c}{ResourceNet} \\ 
{} &             MAPE &  std &            MAPE &  std &                           MAPE &  std &             MAPE &  std  \\
\hline
CPU &     14.9 &     19.6 &                 16.5 &     12.7 &                     14.2 &     12.7 &                   6.4 &     10.4      \\
CPI &     24.4 &     14.9 &                 7.6  &     6.0  &                     7.5  &     5.9  &                   4.1 &     5.9       \\
IOR &     13.7 &     15.6 &                 11.7 &     8.5  &                     10.7 &     8.2  &                   3.8 &     4.5       \\
IOW &     3.5  &     8.0  &                 1.7  &     1.8  &                     1.8  &     1.9  &                   1.4 &     1.6      \\
RAM &     57.7 &     37.6 &                 13.8 &     12.3 &                     12.5 &     12.0 &                   7.0 &     9.1       \\
TLBM &    6.5  &     05.3 &                 3.9  &     3.3  &                     3.9  &     3.3  &                   1.7 &     3.1       \\
PF &      3.7  &     6.9  &                 4.5  &     4.8  &                     4.6  &     4.9  &                   2.7 &     4.4       \\
LLCM &    8.8  &     12.2 &                 6.4  &     7.9  &                     6.4  &     7.6  &                   4.4 &     7.9       \\
FLCM &    8.1  &     08.2 &                 5.3  &     5.3  &                     4.9  &     4.8  &                   3.1 &     4.7       \\ \hline
\end{tabularx}
\end{footnotesize}
\caption{Mean Absolute Percentage Error for each metric and model, evaluated in the test set}
\label{tab:ResultsTable}
\end{table*}

\begin{figure*}[h!]
\centering
  \includegraphics[width=0.6\textwidth]{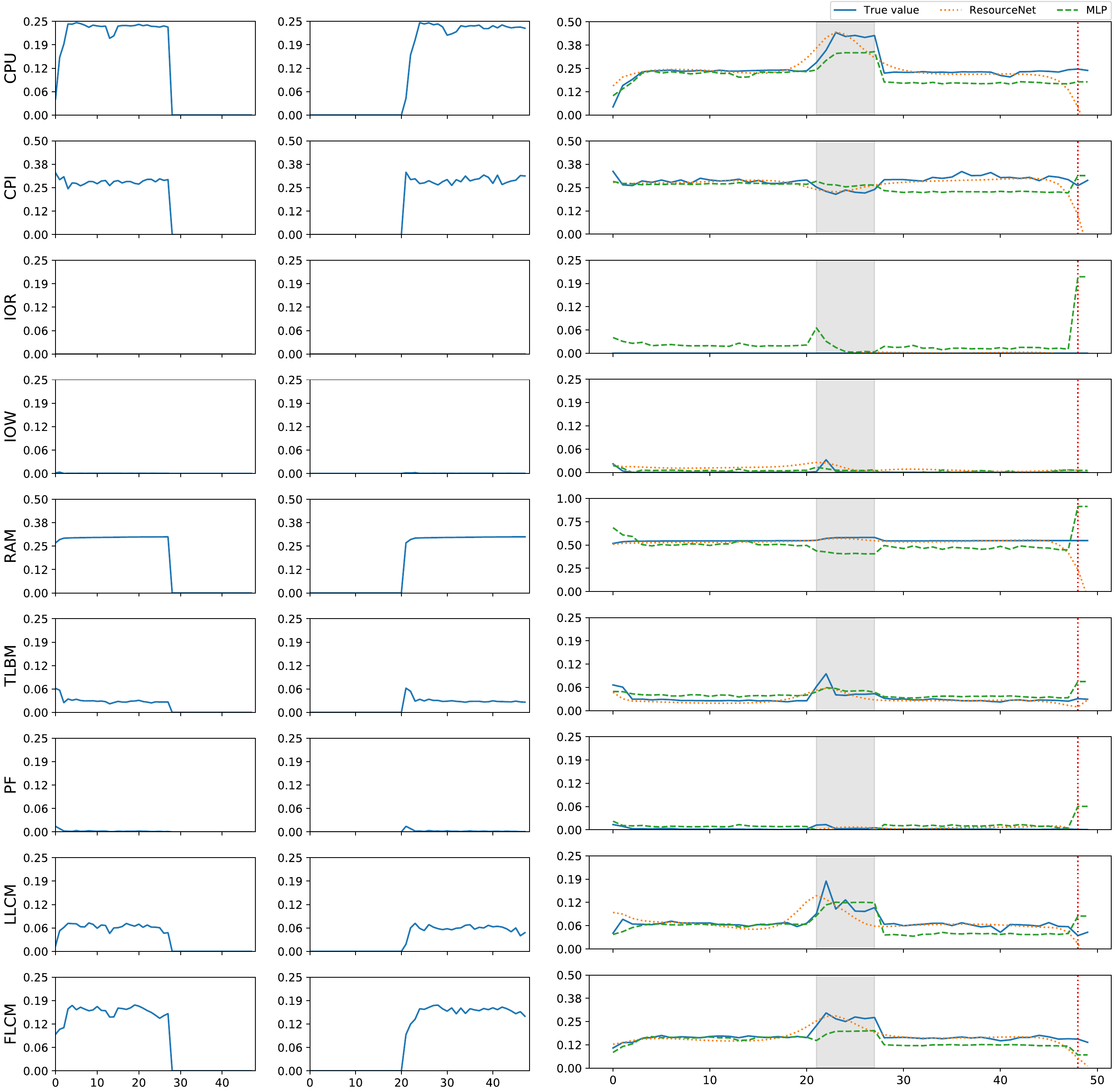}
 \caption{This figure shows two co-located applications that  share  resources only during a fraction of their execution. In this example the second application starts roughly at time step 20  while the first one finishes around time step 25. In this scenario all models can capture the increase in the metrics (specially CPU usage) during the brief period in which both applications run at the same time. Notice that the vertical discontinuous line is around time step 48. This means that the applications took around the same time when they were run in isolation than when they are run co-scheduled.}
 \label{fig:easy_example}
\end{figure*}
\begin{figure*}[h!]
\centering
  \includegraphics[width=0.6\textwidth]{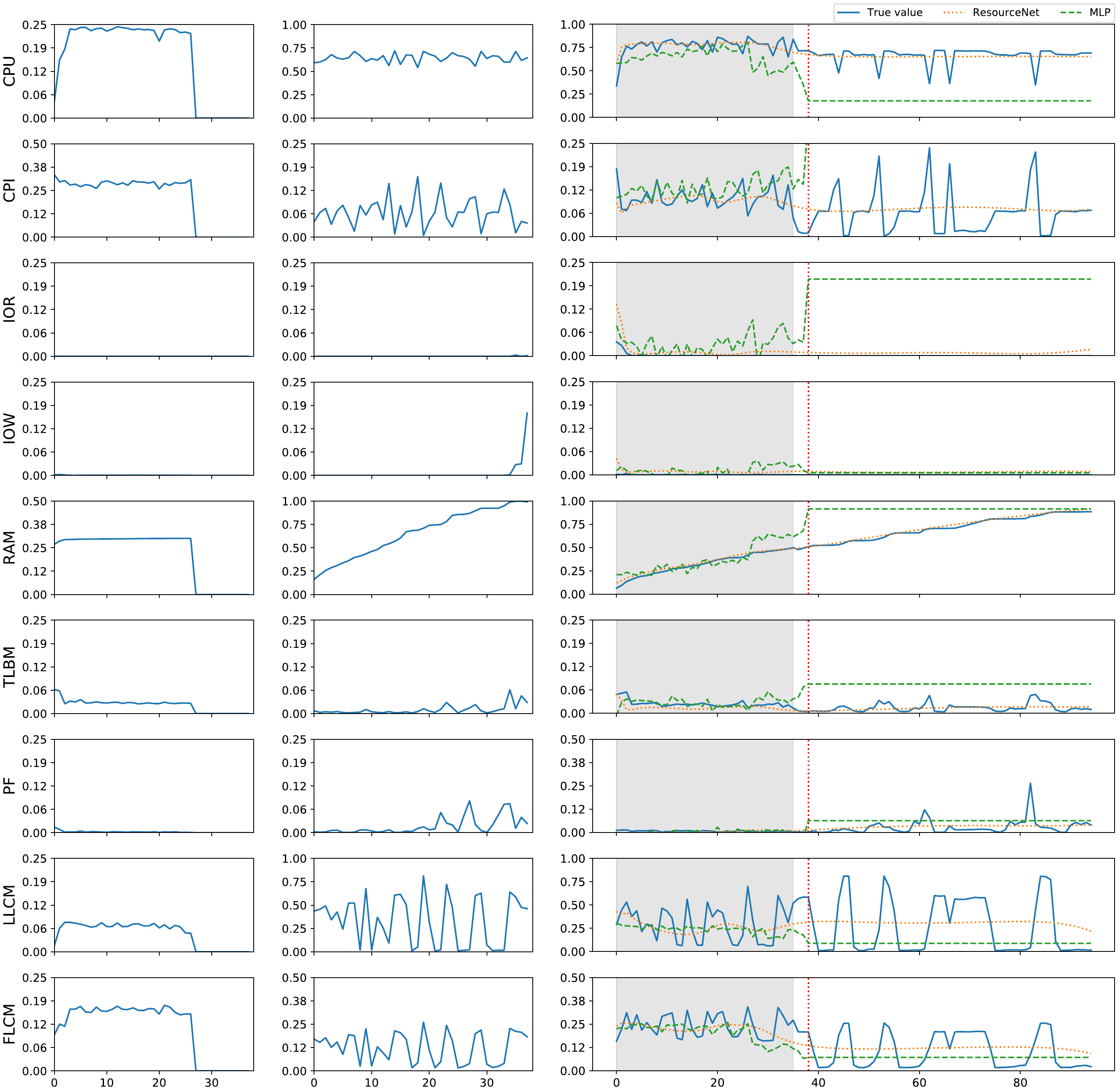}
 \caption{This figure shows a scenario where one application demands almost all memory available. Notice how the MLP makes high error predictions for  CPU and IOR  when the input traces finish (marked by the vertical red line). Nevertheless the resource usage trend is predicted semi-accurately while both applications run. 
 }
\label{fig:Example_test_10}
\end{figure*}
\begin{figure*}[h!]
\centering
 \includegraphics[width=0.6\textwidth]{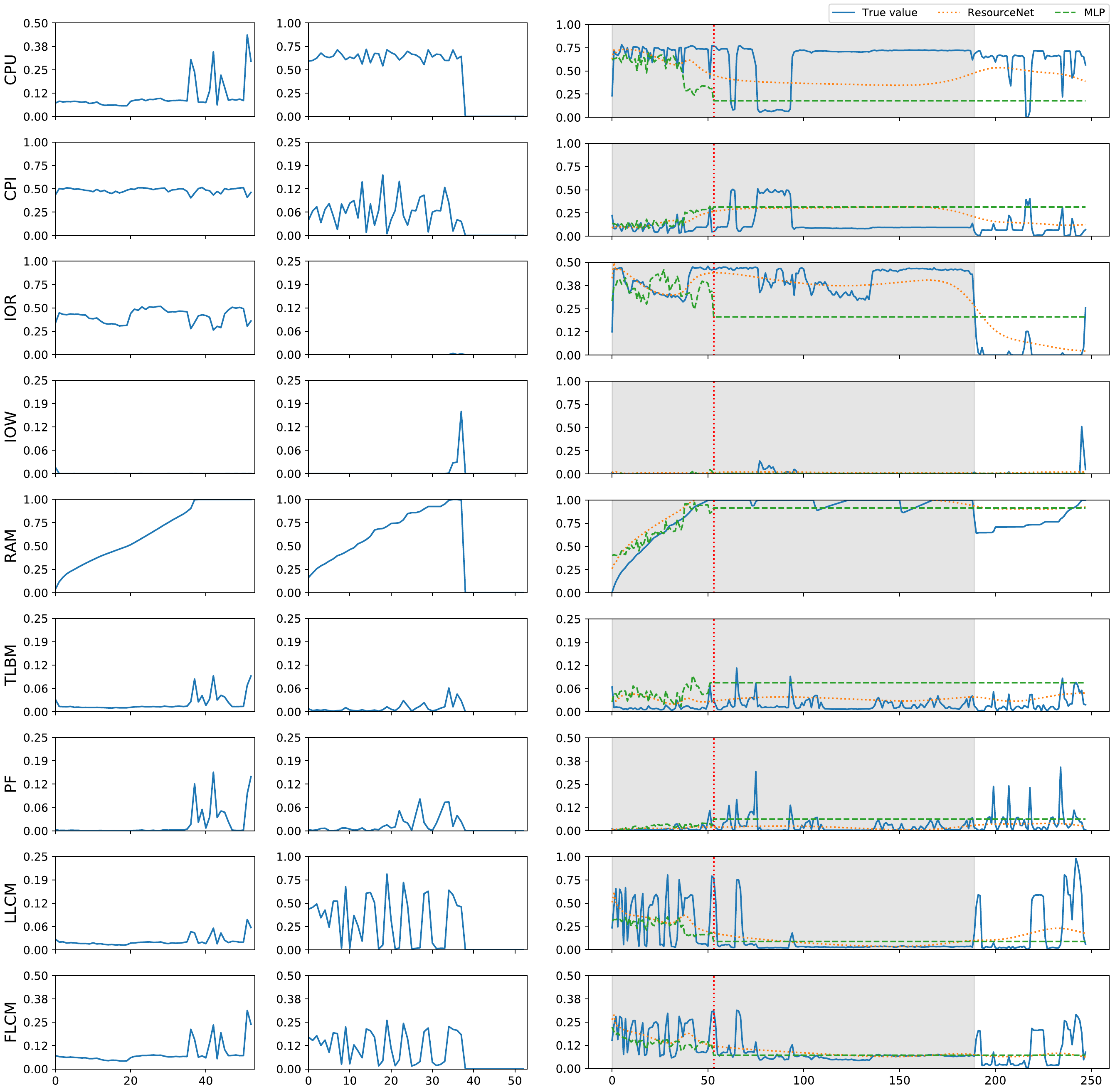}
 \caption{This figure shows a bad co-scheduling scenario where both applications use all the available memory at some point during the execution. Notice that the execution time of  the concurrent applications is roughly five times more the amount they need while running in isolation.}
\label{fig:Example_test_09}
\end{figure*}

To assess the behavior of the presented models in the test set visually, we plotted the resource usage of three triplets. Figures  \ref{fig:easy_example},  \ref{fig:Example_test_10} and \ref{fig:Example_test_09} show three different pairs of co-located applications with different properties. Each figure is composed of three columns. The first two columns show the resources usage trace in the isolated runs. The third column shows the trace of the co-located applications with the predictions made by the MLP and the sequence to sequence model.  The shaded region shown in the third column  displays the period at which  both applications run at the same time. This is the period used to compute the error metrics  (when both applications are running at the same time). Moreover, vertical discontinuous lines mark  the time step at which  the input $\left[ \bm{a};  \bm{b} \right]$ finishes.    In the interest of clarity, the figures do not contain the predictions of the other models in order  to avoid excessively cluttered images.

When applications have low resource demands, and especially when they compete for resources only during small periods of time, the expected demands are easy to predict. Figure~\ref{fig:easy_example} shows a pair of low-demand applications competing for resources during only a small period of time.
In this case, all models are able to predict all metrics well. In contrast, in cases where more demanding applications are co-located, the expected resource usage of the co-located applications over time is not straightforward to predict.  One of the main difficulties in this type of scenarios is the slowdown of both applications while competing for resources, which usually implies a big difference in execution time with respect to the applications being run in isolation. This can be seen in   Figs.~\ref{fig:Example_test_10} and \ref{fig:Example_test_09}. In Fig.~\ref{fig:Example_test_10} the input jobs take around 40 time steps  to execute in isolation, but require more than 80  under the presented co-schedule. The same effect, even more pronounced, can be seen in  Figure~\ref{fig:Example_test_09}. 

In the cases presented, ResourceNet is capable of capturing the trend of the resources correctly; not only in the shadowed region but throughout the entire sequence. For example, the CPU usage in Fig.~\ref{fig:Example_test_10}  starts at around 75\% of usage, then decreases at around 50 and ends again at around 75\% of CPU.  In this figure, one may clearly observe the undesired drawback of the element-wise models previously explained.  Around time step 38, we can see the predictions of the MLP getting stuck at a particular value. This is easy to observe for  IOR, RAM and CPU usage, where predictions are far from the true values. Such behavior is caused due to the lack of input features in the isolated traces, so the model has to rely on making predictions out of zero-padded feature values because either $\bm{a}$ or $\bm{b}$ or both have finished. The sequence-to-sequence model is capable of generating and output sequence longer than the input sequence in a ``natural'' way, without any need to pad the original features.
 
 Figure ~\ref{fig:Example_test_09} shows two jobs which require a high amount of RAM. The first requires low CPU and high IOR, while the second is just the opposite. We can see that once the first job has finished the IOR demand in the co-located trace goes to zero and our model can successfully detect this trend. Even though models capture the trends of all features reasonably well, none of them are able to predict the burstiness of some features, especially LLCM and FLCM.



\subsection{Experiment 2: Estimating co-scheduled execution time}

 \subsection*{Stopping criteria of the co-scheduled trace}
 
 
 
 We have experimented with different criteria  to predict the runtime of  co-scheduled applications. Our methods are based on PC and EOS features which are detailed in   Section \ref{subsection:completion_time}. 
  Using these features, we can test different criteria to predict the completion time of a co-scheduled pair of jobs.
  
  Since we do not know in advance how long co-scheduled applications can take to finish, we have decided to treat the length of the generated trace as a hyperparameter to be adjusted. The different criteria tested are as follows:
 
 \begin{itemize}
 \item \textit{first max EOS}: applications are predicted to finish when the first local maximum is found in $EOS$.
 \item \textit{first max PC sum}:  applications are predicted to finishwhen the first local maximum is found in the sum of $PC_a$ and $PC_b$.
 \item \textit{argmax EOS}:  applications are predicted to finish at the argmax of  the sum of $EOS$.
 \item \textit{argmax PC sum}:   applications are predicted to finish at the argmax of  the sum of $PC_a$ and $PC_b$.
 \end{itemize}


For each of the criteria,  Table \ref{tab:ResultsTableStopping1} reports the MAPE error and the standard deviation depending on the length of the generated trace. The first column \textit{length}  contains at position $k x$ the expected error when the co-scheduled trace is generated  up to $k$ times $x$ timesteps, where $x$ is the longest of the input traces  when executed in an isolated environment.  Results show than if the length of the generated trace is too short (for example $1x$),  then errors are large because the stopping criteria is fired before the co-scheduled applications might finish. Errors decrease as $k$ increases up to a point where the errors start to increase. The best results are achieved at $k=4$.  Our experiments show much lower errors  for the criteria based on PC features than for criteria based on the EOS feature.
\begin{table}[h!]
\centering
\begin{scriptsize}
\begin{tabular}{lrrrrrrrr}
\toprule
{} & \multicolumn{2}{l}{first max EOS} & \multicolumn{2}{l}{first max PC sum} & \multicolumn{2}{l}{argmax EOS} & \multicolumn{2}{l}{argmax PC sum} \\
length &          MAPE &  std &             MAPE &   std &       MAPE &   std &          MAPE &   std \\
\midrule
1x &          96.6 &  5.1 &             98.3 &   8.1 &       77.5 &  21.3 &          59.0 &  17.7 \\
2x &          96.6 &  5.1 &             77.7 &  39.1 &       49.1 &  35.9 &          25.1 &  20.4 \\
3x &          96.6 &  5.1 &             46.2 &  45.2 &       18.8 &  24.7 &          14.3 &  15.6 \\
4x &          96.6 &  5.1 &             29.4 &  37.8 &       30.8 &  35.9 &          12.7 &  18.2 \\
5x &          96.6 &  5.1 &             29.4 &  37.8 &       31.5 &  45.6 &          28.2 &  39.0 \\
6x &          96.6 &  5.1 &             29.4 &  37.8 &       43.8 &  64.5 &          30.0 &  44.1 \\
7x &          96.6 &  5.1 &             28.1 &  37.0 &       48.8 &  75.1 &         106.9 &  96.9 \\
8x &          96.6 &  5.1 &             28.1 &  37.0 &       63.6 &  97.2 &         107.6 &  98.6 \\
\bottomrule
\end{tabular}
\end{scriptsize}
 \caption{MAPE errors made by predicting the execution time of co-scheduled applications  for the different criteria based  EOS and PC features.}
 \label{tab:ResultsTableStopping1}
\end{table}

\begin{figure}[h!]
\centering
 \includegraphics[width=0.8\linewidth]{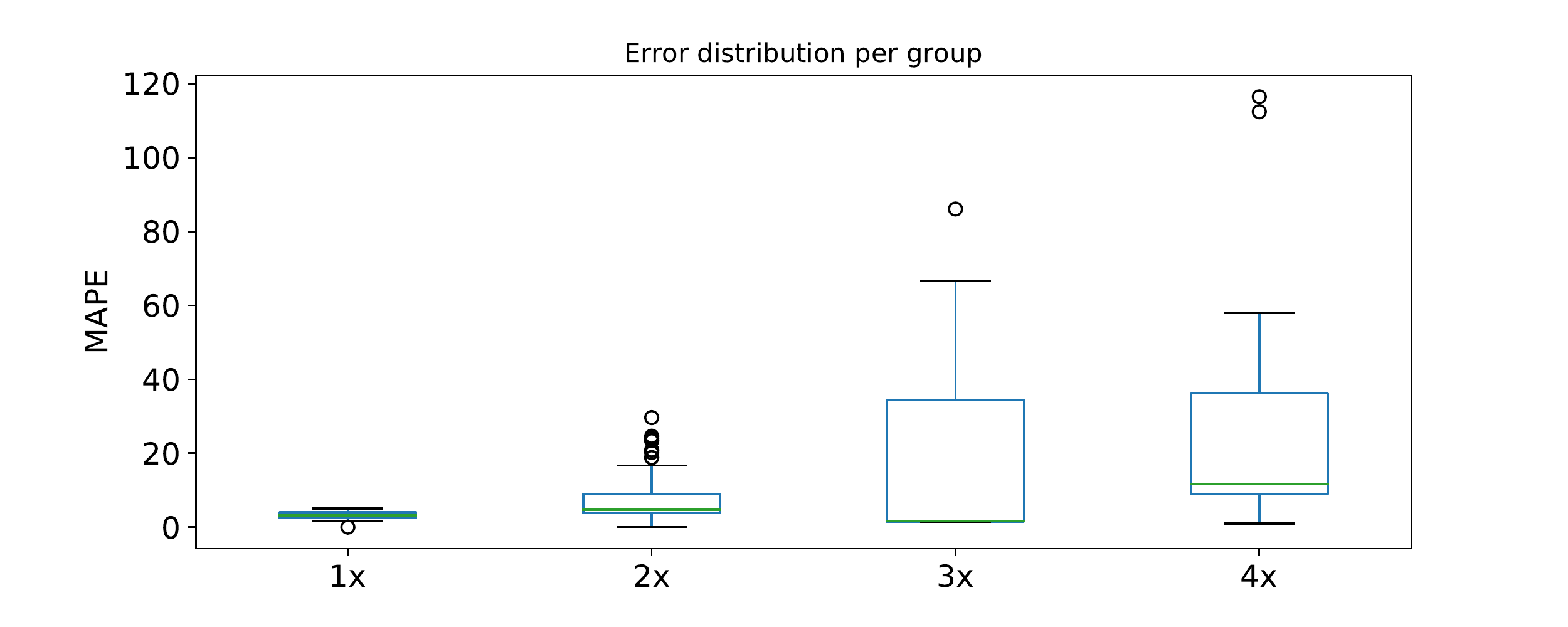}
 \caption{This figure shows the distribution of the MAPE grouped into 4 categories. The farther away the co-located applications are predicted to finish the larger the expected error. The box plot shows results for the criterion \textit{argmax PC sum}. }
 \label{fig:binned_results}
\end{figure}

The errors made by our criteria behave differently in accordance with the length of time we predict applications will need to finish with respect to the runs made in isolation.  In Fig~\ref{fig:binned_results} one may observe the distribution of errors made by our criteria, grouped according to where  we predict the execution will end.  Applications that are expected to run between $1x$ and $2x$ have low errors, while applications that are expected to finish between $3x$ and $4x$ have higher errors and more variance. 
Notice that this knowledge can be used as a rule for deciding when applications should not be co-scheduled. A simple rule for avoiding co-scheduling applications could be to forbid scenarios in which the model runtime prediction falls within $3x$ or $4x$.

 \subsection*{Looking at EOS and PC feature predictions}

\begin{figure*}[h!]
\centering
 \includegraphics[width=0.99\textwidth]{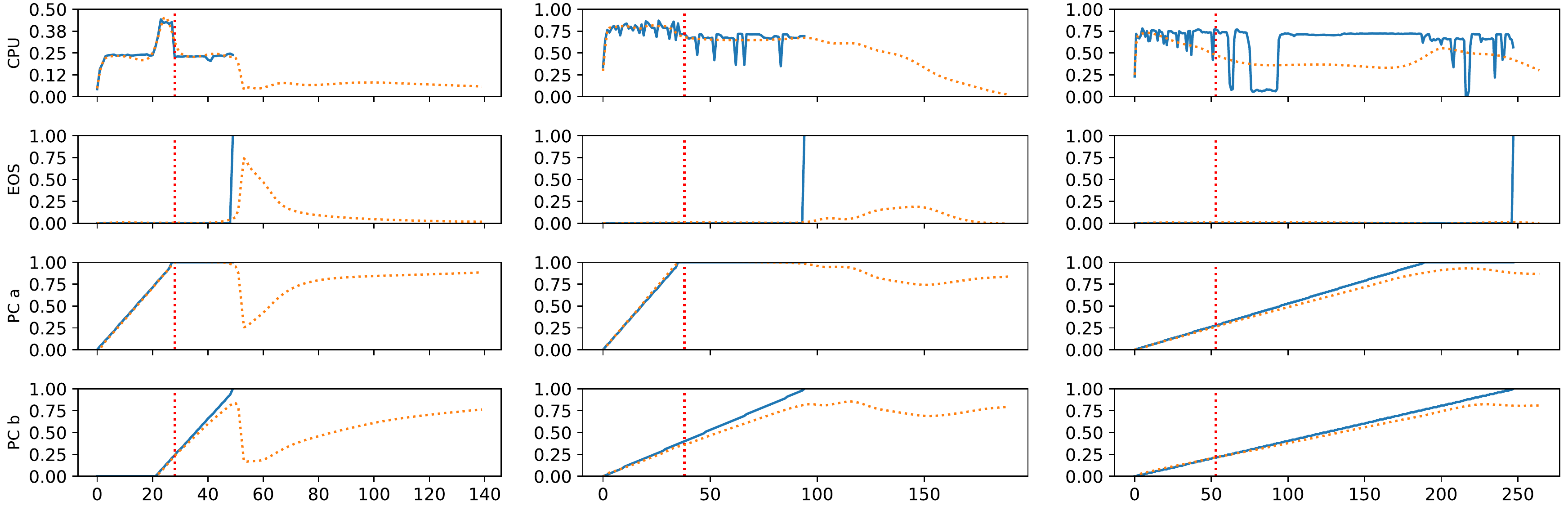}
 \caption{This figure shows the EOS and PC features for the workloads seen in Figures [\ref{fig:easy_example}]  [\ref{fig:Example_test_10}] and [\ref{fig:Example_test_09}]. Every column corresponds to a  different colocation of two workloads. The continuous lines show the true values of the different features  and  the dotted traces show the predictions made by the sequence to sequence model. Vertical lines show the length of the longest workload executed in isolation. Plots in this figure show predictions generated up to five times the length of the longest input sequence. The CPU resource usage of the co-scheduled pair has been added into the plot to contextualize the other features.  }
 \label{fig:end_o_sequence}
\end{figure*}

Figure \ref{fig:end_o_sequence} shows the EOS and PC features behaviour for three different co-scheduled workload pairs. Notice that the longer a pair takes to execute, the less a signal is shown by the EOS feature. Nevertheless, PC features maintain a relatively good quality even for co-scheduled pairs with high resource competition. 

Our  results suggest  that criteria  build to  predict the length of a co-scheduled trace using EOS should not be based on the actual value of the EOS feature, but rather on a function taking into account the maximum value achieved during the prediction process. Notice that, in the first case, the predicted EOS arrives roughly at 0.75. However, the second case barely takes value
0.25 and the third case the EOS scarcely barely goes above zero. This behavior increases the difficulty of predicting runtime even more with this feature.
 
 A reasonable explanation for the behavior of EOS is that our model is trained by minimizing the mean squared error be- tween predictions and true traces. It is logical to argue that the optimization procedure induces the network to focus more on predicting PC features better than EOS, since the EOS feature takes value 1 at only 1 time step while PC features increase at every time step.  The overall contribution to the error of always predicting 0 in EOS is therefore much lower than always predicting 0 in $PC_a$ or $PC_b$.

\section{Related Work}
\label{sec:relatedwork}

This section presents related work that is relevant for resource, runtime and interference prediction. The three topics are closely related and still an active field of research with many different applications. Works that target those topics usually take into account more than one topic since they are closely related. Two relevant applications are dynamic provisioning and job placement. In dynamic provisioning scenarios the main goal is to predict whether there will be a change in the workload trace that needs some hardware decision to take place (like pro- visioning more nodes for a workload). In the case of job place- ment, resource predictions also play an important role, since similar resource needs for different jobs resources may affect. Thus, schedulers may aim for colocations of applications that do not degrade when trying to maximize the use of available hardware.

\subsection*{Resource prediction related work}

The main goal of oracles in workload provisioning scenarios is to guess accurately when workloads will need more resources in the future. Works like \cite{Islam2012} and \cite{Kumar2018} use neural networks that take as input resource usage in a given time window the aim of predicting the future resource requirements for the workloads. In \cite{Kumar2018} authors use differential evolution as a means to train the models, whereas in \cite{Islam2012}  they use standard gradient based algorithms.

Works like Sanz-Marco et al.~\cite{Marco2017} apply machine learning to forecast memory requirements towards interference avoidance. 
They proposed a mixture of expert approach to predict the amount of memory needed by a SPARK application, given the size of the input data. From a dataset consisting of applications and SPARK profiling features, captured from hardware counters (CPI, number of interruptions, number of cache misses, etc). Then, after using Principal Component Analysis to reduce dimensionality, the mixture of experts each learn regression function to predict the required memory using the size of input data as input. The application is first profiled with a small input 10\%, and it then passes through a k-Nearest Neighbor in order to choose the most representative expert. Finally, the application is run with different data sizes in order to tune the function parameters to determine the best fit for this application.

\subsection*{Interference focus-related work}


In ~\cite{Delimitrou2013}  the authors present Paragon, an approach to predicting the interference between two applications by modeling them using SVD and PQ-reconstruction~\cite{SVD}.
In Paragon, the profiling across pairs is limited to the first minute due to time constraints, without covering applications with different execution phases and behaviors over time, although they acknowledge 
the problem by trying to mitigate it by labeling nodes with unexpected changes in resource demands, then leave them running until jobs are finished.


Heracles et al.~\cite{Lo2015} present an approach for interference mitigation in high priority applications through job isolation techniques. Among other tools, their  solution uses \textit{cpugroups}, \textit{qDisc} or the \textit{CAT} technology~\cite{CAT}  to properly ensure resource availability for specific jobs. 
Such a solution tends to over- resource applications for solving the interference problem, thus becoming subject to machine under-utilization. Furthermore, the optimization method poses an NP-hard solution, which makes it feasible for only a small number of jobs.

In a similar line, Mishra et al.~\cite{Mishra2017} presented ESP, a predictive model for forecasting the interference between two applications. ESP uses low level metrics as input for the model, but instead of making predictions at sequence level they are held at a scalar level, both at the input and the output, by means of aggregation metrics such as the average of the sequence. Thus, given the average of two metrics, it predicts the average of the concurrent execution in a two-step model as well; one for feature selection and the other for using a regression model to retrieve the expected aggregated metric as output when both tasks are co-located.

Another interesting work dealing with protection against interference is Stay-Away by ~\cite{Rameshan2014}. The idea they propose starts by projecting the resource consumption of the applications in a 2-dimensional space. They then predict how applications will move in that 2D space. If applications are predicted to have collisions between applications in that space, they then adopt a counteraction.


\subsection*{Run-time prediction-related work}

In \cite{Wyatt2018} the authors present a convolutional neural network named PRIONN. The model  predicts run-time and IO (bytes read and total bytes written) of applications based on the source code in the input script.  This work maps job scripts code into image-like representations so there is no need for a feature generation process to process the raw text of the scripts.

In \cite{DBLP:conf/kdd/BerralPCCRG15} Aloja-ML is presented as a framework for characterization and knowledge discovery in Hadoop deployments. The authors present different methods for execution time prediction based on hardware resources, workload type and input data size.

In \cite{Wang2017} a mixture of runtime prediction and interference is presented in two methods: The first, aimed at calculating the cost of a Spark job run in isolation, is approximated by modeling a function from the execution parameters (the number of data partitions, the number of stages, the number of jobs per stage, etc.) to predict the total cost of the job. The second, designed to predict the cost when two jobs run in concurrency by modeling interference, is tackled by adapting the previous formulas. They run a small part of the input data for each Spark stage combination to compute a ration of interference $\beta$, to be added to the modeling function as a factor of the addition of required resources as $resource_i(AB) = \beta \cdot (resource_i(A) + resource_i(B))$. This work recognizes the different behavior across the execution caused by the different phases of executions, but is limited to predefined Spark phases that have to be given by the user.


\section{Conclusions }
\label{sec:conclusions}

 This paper introduces the use of recurrent neural networks for interference and resource prediction tasks of co-scheduled applications. We adapt a sequence-to-sequence model to work with two input sequences instead of one, thereby providing a padding containing the start time of one application with respect to the other. Our method predicts resource usage of the monitored metrics over time and is adaptable enough to be trained with workloads of arbitrary input and output lengths. Moreover, since training is done for a regression task instead of a classification task, we introduce the percentage completion features which significantly improve completion time prediction of co-scheduled applications.

Our experiments show that the model is able to predict the resource usage of co-located applications sharing resources over time, even when application performance degrades drastically due to a high demand of similar resources at the same time. Moreover, as a baseline, we compare our model with standard machine learning algorithms. The experiments show that our model makes more accurate predictions and is able to deal with the sequential nature of the data, thus making it suitable for the presented scenario where input/output pairs have different lengths.

\section*{Funding }

This work is supported by the European Research Council (ERC) under the European Union's Horizon 2020 research and innovation programme (grant agreement no. 639595); de Catalunya under contract 2014SGR1051; the ICREA Academia program; and the BSC-CNS Severo Ochoa program (SEV-2015-0493); the Spanish Ministry of Economy under contract TIN2015-65316-P and the Generalitat.

\section{Acknowledgements}

Funding: This work is supported by the
European Research Council (ERC) under the European Union's Horizon 2020 research
and innovation programme (grant agreements nº 639595);
the Spanish Ministry of Economy under contract TIN2015-65316-P and the Generalitat
de Catalunya under contract 2014SGR1051;
the ICREA Academia program;
and the BSC-CNS Severo Ochoa program (SEV-2015-0493);




\authorbibliography[scale=0.15,wraplines=10,overhang=40pt,imagewidth=4cm,imagepos=r]{dbuchaca.jpeg}{David Buchaca Prats}{He recieved his degree in Mathematics in 2012 and a M.Sc in Artificial Intelligence in 2014. He is an applied mathematician, working in applications of Artificial Neural Networks. He is currently a Ph.D. student at the Barcelona Supercomputing Center within the ``Data-Centric Computing'' research line.}

\authorbibliography[scale=0.2,wraplines=10,overhang=40pt,imagewidth=4cm,imagepos=r]{jmarcual.jpeg} {Joan Marcual}{He acquired a degree in Computer Science in 2017 from the Polytechnic University of Catalonia. He is currently pursuing a M.Sc in Data Science. His main research interest is machine learning applied to health data.}

\authorbibliography[scale=0.3,wraplines=10,overhang=40pt,imagewidth=4cm,imagepos=r]{jberral.jpeg}{Josep Llu\'is Berral}{He received his degree in Informatics (2007), M.Sc in Computer Architecture (2008), and Ph.D. at BarcelonaTech-UPC (2013). He is a data scientist, working in applications of data mining and machine learning on data-center and cloud environments at the Barcelona Supercomputing Center (BSC) within the ``Data-Centric Computing'' research line. He has worked at the High Performance Computing group at the Computer Architecture Department-UPC, also at the Relational Algorithms, Complexity and Learning group at the Computer Science Department-UPC. He received in 2017 a Juan de la Cierva research fellowship by the Spanish Ministry of Economy. He is an IEEE and ACM member.}

\authorbibliography[scale=0.3,wraplines=10,overhang=40pt,imagewidth=4cm,imagepos=r]{dcarrera.jpeg}{David Carrera}{He received the MS degree at BarcelonaTech-UPC in 2002 and his PhD from the same university in 2008. He is an associate professor at the Computer Architecture Department of the UPC. He is also an associate researcher in the Barcelona Supercomputing Center (BSC) within the “Data-Centric Computing” research line. His research interests are focused on the performance management of data center workloads. He has been involved in several EU and industrial research projects. In 2015 he was awarded an ERC Starting Grant for the project HiEST. He received an IBM Faculty Award in 2010. He is an IEEE member.}

\bibliographystyle{elsarticle-num}
\bibliography{mybib.bib}

\end{document}